\documentclass[]{spie}  

\usepackage{amsmath,amsfonts,amssymb}
\usepackage{graphicx}
\usepackage[colorlinks=true, allcolors=blue]{hyperref}

\title{Optimizing contrastive learning for cortical folding pattern detection}

\author[a]{Aymeric Gaudin}
\author[a]{Louise Guillon}
\author[a]{Clara Fischer}
\author[b]{Arnaud Cachia}
\author[a]{Denis Rivière}
\author[a]{Jean-François Mangin}
\author[a]{Joël Chavas}
\affil[a]{Neurospin, Bâtiment 145, CEA Saclay, 91191 Gif-sur-Yvette, France}
\affil[b]{LaPsyDé, Laboratoire A.Binet-Sorbonne, 46 rue Saint Jacques, 75005 Paris}

\begin{document} 
\maketitle

\begin{abstract}
The human cerebral cortex has many bumps and grooves called gyri and sulci. Even though there is a high inter-individual consistency for the main cortical folds, this is not the case when we examine the exact shapes and details of the folding patterns. Because of this complexity, characterizing the cortical folding variability and relating them to subjects' behavioral characteristics or pathologies is still an open scientific problem. Classical approaches include labeling a few specific patterns, either manually or semi-automatically, based on geometric distances, but the recent availability of MRI image datasets of tens of thousands of subjects makes modern deep-learning techniques particularly attractive. Here, we build a self-supervised deep-learning model to detect folding patterns in the cingulate region. We train a contrastive self-supervised model (SimCLR) on both Human Connectome Project (1101 subjects) and UKBioBank (21070 subjects) datasets with topological-based augmentations on the cortical skeletons, which are topological objects that capture the shape of the folds. We explore several backbone architectures (convolutional network, DenseNet, and PointNet) for the SimCLR. For evaluation and testing, we perform a linear classification task on a database manually labeled for the presence of the "double-parallel" folding pattern in the cingulate region, which is related to schizophrenia characteristics. The best model, giving a test AUC of 0.76, is a convolutional network with 6 layers, a 10-dimensional latent space, a linear projection head, and using the branch-clipping augmentation. This is the first time that a self-supervised deep learning model has been applied to cortical skeletons on such a large dataset and quantitatively evaluated. We can now envisage the next step: applying it to other brain regions to detect other biomarkers. The GitHub repository is publicly available on  \href{https://github.com/neurospin-projects/2022_jchavas_cingulate_inhibitory_control}{https://github.com/neurospin-projects/2022\_jchavas\_cingulate\_inhibitory\_control}.

\end{abstract}
%

\section{Introduction}

The human cerebral cortex is folded and made of gyri separated by folds. Even though the main folding patterns are common across humans, the details of the cortical folding present a very high inter-subject variability, just like fingerprints. This variability is both a challenge and an opportunity: studying it can let us find anatomical biomarkers of neurodevelopmental pathologies or of subjects' endophenotypes linked to genetic polymorphisms.

We can decipher this variability by looking for local shape features called cortical folding patterns or folding patterns.

\begin{figure}
\centering
\includegraphics[width=0.9\textwidth]{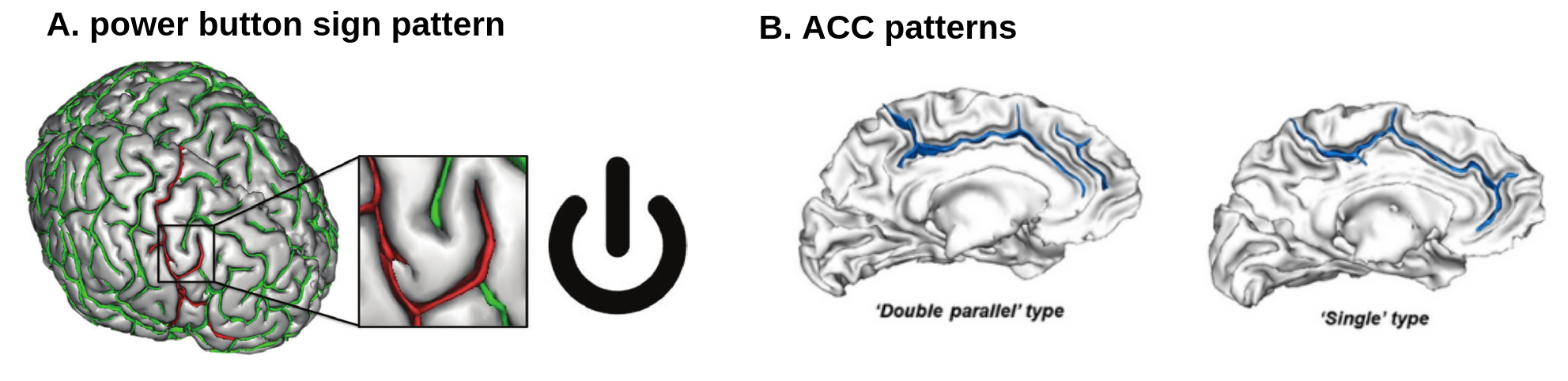}
\caption{\textit{Examples of cortical patterns.} \textbf{A.} The power button sign pattern in the central region is linked to epilepsy~\cite{mellerio_power_2015}. \textbf{B.} The anterior cingulate cortex (ACC) has two known patterns, a double-parallel pattern and a single-fold pattern~\cite{cachia_shape_2014}. We use the detection of these two ACC patterns to evaluate our models.} \label{fig:examples}
\end{figure}

The literature shows few cases of cortical folding patterns that have been identified and correlated with pathologies or functions. For example, in the precentral region, Mellerio et al. (2015) manually identified a pattern named the Power Button Sign (PBS) pattern because of its recognizable shape (Fig.~\ref{fig:examples}A). This pattern is very rare in the general population but is present in about 60\% of patients with dysplasia near the motor area, leading to epilepsy~\cite{mellerio_power_2015}. Also, the Anterior Cingulate Cortex (ACC) has two identified folding patterns, the single-fold pattern and the double-parallel pattern (Fig.~\ref{fig:examples}B)~\cite{yucel_hemispheric_2001}: the presence of different ACC patterns in both hemispheres is related to control efficiency in preschoolers~\cite{cachia_shape_2014}, and the shape of the ACC folds has been related to hallucinations among human patients suffering from schizophrenia~\cite{garrison_paracingulate_2015}. To complement the manual methods, Borne et al. proposed a supervised model based on a convolutional network to detect known patterns~\cite{borne_automatic_2021}. 

These manual techniques have a drawback: they are long to implement, while the supervised ones will not permit the detection of new folding patterns. We propose to resort to unsupervised techniques that allow the use of the large datasets that are now available. Recently, $\beta$-VAE~\cite{higgins_beta-vae_2017} and SimCLR~\cite{chen_simple_2020} models have been trained to build a latent space for folding representation~\cite{chavas_unsupervised_2022,guillon_detection_2021,guillon_identification_2024}. However, these studies have not used a manually labeled dataset for hyperparameter tuning and evaluation.

Here, we build a contrastive learning framework to find a representation that permits the automatic detection of relevant folding patterns. We use as inputs the so-called cortical skeletons---topologically-defined surfaces that follow the middle of a sulcus (Fig~\ref{fig:representations}, blue surface in the left image)---, which are stable throughout life after early childhood and are expected to be less sensitive to data acquisition sites. We optimize a SimCLR model and evaluate it by detecting the double-parallel pattern in the anterior cingulate cortex of the right hemisphere.

\section{Methods}

\subsection{Datasets}\label{method:dataset}

\begin{figure}[h]
\centering
\includegraphics[width=0.7\textwidth]{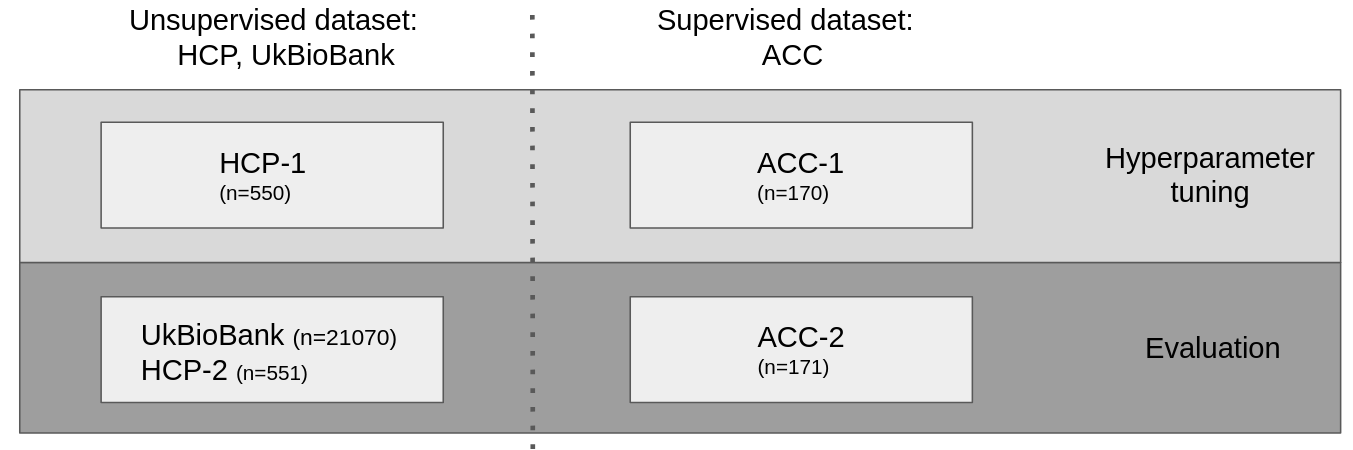}
\caption{\textit{Dataset splits used for parameter tuning and evaluation.}}\label{fig:split}
\end{figure}

Three datasets are used in this project: the human connectome project (HCP) dataset, the UKBioBank dataset, and the so-called ACC dataset~:
\begin{itemize}
\item Data collection and sharing for the HCP dataset was provided by the MGH-USC Human Connectome Project~\cite{van_essen_wu-minn_2013}. We used in this work MRIs of 1101 HCP subjects,
\item We preprocessed 21070 subjects of the UKBioBank, which recruited 500,000 people~\cite{sudlow_uk_2015},
\item The ACC dataset stands for the Anterior Cingulate Cortex dataset. This dataset contains MRIs from 341 young subjects taken from 4 separate studies~\cite{chakravarty_striatal_2015,rapoport_childhood_2011,cachia_longitudinal_2016,delalande_complex_2020,tissier_sulcal_2018}. Each MRI has been manually labeled for the presence or absence of the paracingulate fold, which lies parallel above the anterior cingulate fold~\cite{ono_atlas_1990,yucel_hemispheric_2001}. The presence of the paracingulate fold corresponds to what we call in this paper the double-parallel pattern.
\end{itemize}
The HCP dataset is randomly split in two, HCP-1 and HCP-2. The ACC dataset is also randomly split in two, ACC-1 for parameter tuning and ACC-2 for testing, using a stratification guaranteeing equipartition of the sites (as there can be an effect of the site on MRI data acquisition), gender, and subjects having a paracingulate fold in the right hemisphere (Fig.~\ref{fig:split}).

\subsection{Demographic information}
The 341 subjects of the ACC dataset are distributed as follows: 197 subjects are from a longitudinal study of childhood-onset schizophrenia~\cite{chakravarty_striatal_2015,rapoport_childhood_2011}, out of which 73 are schizophrenia patients, 49 are patients' siblings, and 75 are healthy subjects~\cite{cachia_longitudinal_2016}. 65 subjects (46 adolescents, 19 children) are taken from Delalande et al., 2020~\cite{delalande_complex_2020}. Last, 79 subjects are taken from Tissier et al., 2018~\cite{tissier_sulcal_2018}. All 341 subjects have been labeled manually for the presence of the paracingulate fold.

Gender distribution is given in Table.~1. HCP subjects were recruited from Missouri (United States), UKBioBank subjects in 2006-2010 from across the UK, whereas ACC subjects were from France and the USA (Table~1).

\begin{table}[h]
\begin{center}
\resizebox{0.8\textwidth}{!}{%
\begin{tabular}{c|ccccc}
    & \#subjects & Age                  & \% of male & Origin      & \#sites \\
    &            & Mean {[}min, max{]} &            &             &         \\
\hline 
UKB & 21070      & 64 {[}44, 82{]}      & 47         & UK          & 1       \\
HCP & 1101       & 29 {[}22, 40{]}      & 48         & USA         & 1       \\
ACC & 341        & 15 {[}8, 40{]}       & 58         & France, USA & 4      
\end{tabular}%
}
\vspace{0.3cm}
\label{tab:demography}
\caption{Demographic information of used datasets.}
\end{center}
\end{table}

\subsection{Preprocessing}\label{method:preprocessing}

\begin{figure}
\centering
\includegraphics[width=0.75\textwidth]{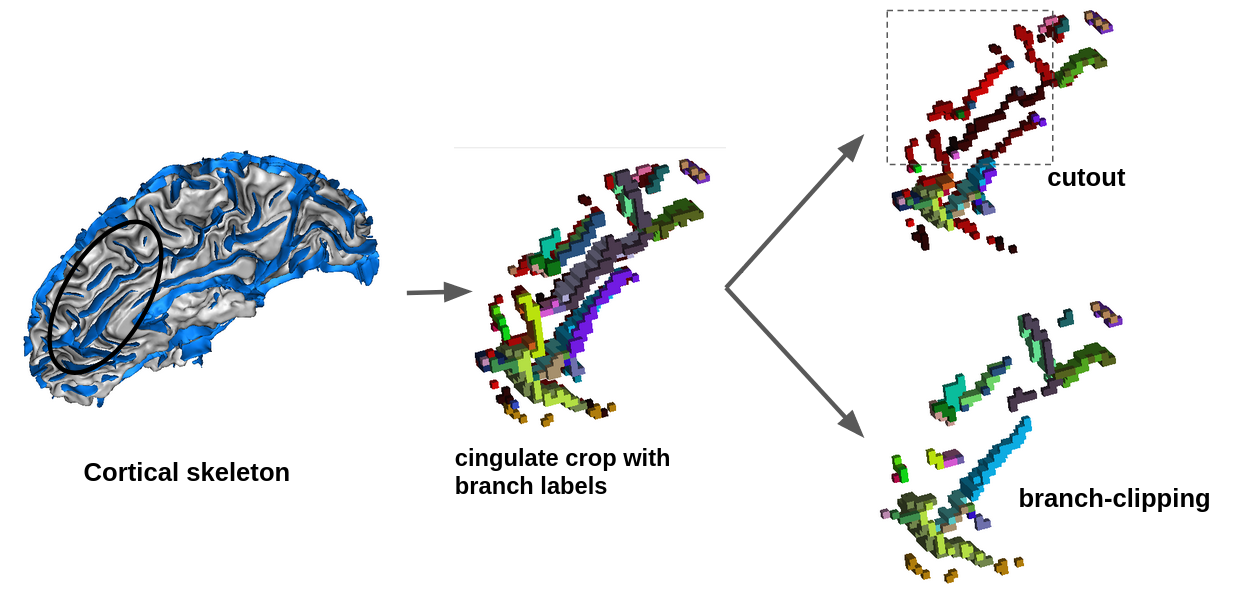}
\caption{\textit{Preprocessing the inputs of the deep learning algorithm.} \textbf{Left.} The cortical skeleton (in blue) represents the cortical folds for the right hemisphere. \textbf{Middle.} The cingulate crop with branch labels (one color per branch label): each branch represents a simple surface, a junction line, or a bottom line of a fold. \textbf{Right.} From the cingulate crop, we construct random views, using either the cutout augmentation, in which only bottom branches (represented in red) are kept inside the cutout (top right), or the branch-clipping augmentation, in which branches are randomly removed, and all bottom branches are removed (bottom right) .} \label{fig:representations}
\end{figure}

All subjects' brain magnetic resonance images (MRIs), affinely normalized to a standard brain referential (ICBMc2009),  were preprocessed with the Morphologist pipeline from the BrainVISA software\footnote{\url{https://brainvisa.info}} to get their folding graph (Fig.~\ref{fig:representations}), namely a graph-based representation of the cortical skeleton.  Each branch of the graph is either a simple surface, a junction line between simple surfaces, or a bottom line of a fold; they are connected in the graph if their corresponding surfaces touch each other~\cite{mangin_3d_1995,riviere_automatic_2002}.

We focus our study on the cingulate region of the right hemisphere (Fig.~\ref{fig:representations}). For this, we compute a mask of the cingulate region over a database where the folds were manually labeled~\cite{borne_automatic_2020}. Our final inputs are 2-mm resolution 3D crops of dimension 17x40x38 (Fig.~\ref{fig:representations}, middle).

\subsection{Model}
The contrastive model used is SimCLR, an instance discrimination contrastive model~\cite{chen_simple_2020}. For each input image $x$ of a batch of size N, we generate two views, $x_i$ and $x_j$, at each epoch, whose model outputs are $z_i$ and $z_j$, respectively. The model trains to bring together views from the same image and moves away views of different images, in that it minimizes $\sum_{i=1}^{N} \ell_{i,j=pos(i)}+\sum_{j=1}^{N} \ell_{j,i=pos(j)}$, with~:
\begin{equation}
\label{eq:loss}
    \ell_{i,j} = -\log \frac{\exp(\mathrm{sim}(z_i, z_j)/\tau)}{\sum_{k=1, k \neq i}^{2N}\exp(\mathrm{sim}(z_i, z_k)/\tau)}
\end{equation}

$pos(i)$ represents the positive pair associated with i, $\tau$ is a temperature parameter, and sim(.,.) is the cosine similarity function.

\subsection{Topology and graph-based augmentations}\label{method:augmentations}

We test two different sets of topology-based transformations to generate the SimCLR views (see~\S\ref{method:optim} for details on the parameters optimization and the right panel of Fig~\ref{fig:representations} for illustration). The first one, called "cutout",  produces two different views: in one view, we remove a rolling block whose dimensions are 55\% of the image volume dimensions. In the second view, all non-zero voxels inside a random 3D block of the same size are kept, and all non-zero voxels outside this block are removed~\cite{chavas_unsupervised_2022}. In all views, fold bottom pixels are conserved to keep the information on the global longitudinal shape of the sulcus.

The second augmentation method is called "branch-clipping": for both views, it removes random branches from the folding graph until there are at least 40\% of the voxels removed. All bottom voxels are removed.

We then binarize the image for all views and apply a random rotation with nearest-neighbor interpolation with a maximum amplitude of 6° around each axis.

\subsection{Backbones}

We tested three different backbones. The first is a simple six-layer convolutional network, taken from a $\beta$-VAE encoder optimized for cortical skeleton images~\cite{chavas_unsupervised_2022}. The second one is a DenseNet architecture~\cite{huang_densely_2018}, which gives good results when used on brain MRI data~\cite{dufumier_contrastive_2021}. Its number of dense blocks is reduced to two, considering the smaller image size~\cite{chavas_unsupervised_2022}. 

The third is a PointNet~\cite{charles_pointnet_2017}. A PointNet is a neural network that, instead of taking a binary 3D matrix with a fixed size as input, takes point clouds, i.e., a list of 3D coordinates corresponding to the non-zero pixels in the original 3D matrix. Indeed, in our volumes, only 4\% of the pixels have non-zero values, making this backbone attractive. We take the original implementation of PointNet in this work. 

\subsection{Linear classification}

A standard way to assess the quality of a SimCLR model is to train classifiers on the embeddings it generates~\cite{chen_simple_2020}. We are using the scikit-learn implementation of a linear Support Vector Classifier (SVC) with probability estimates: we take as a criterion the area under the receiver operating characteristic curve (AUC)~\cite{pedregosa_scikit-learn_2011}. 

\begin{figure}
\centering
\includegraphics[width=\textwidth]{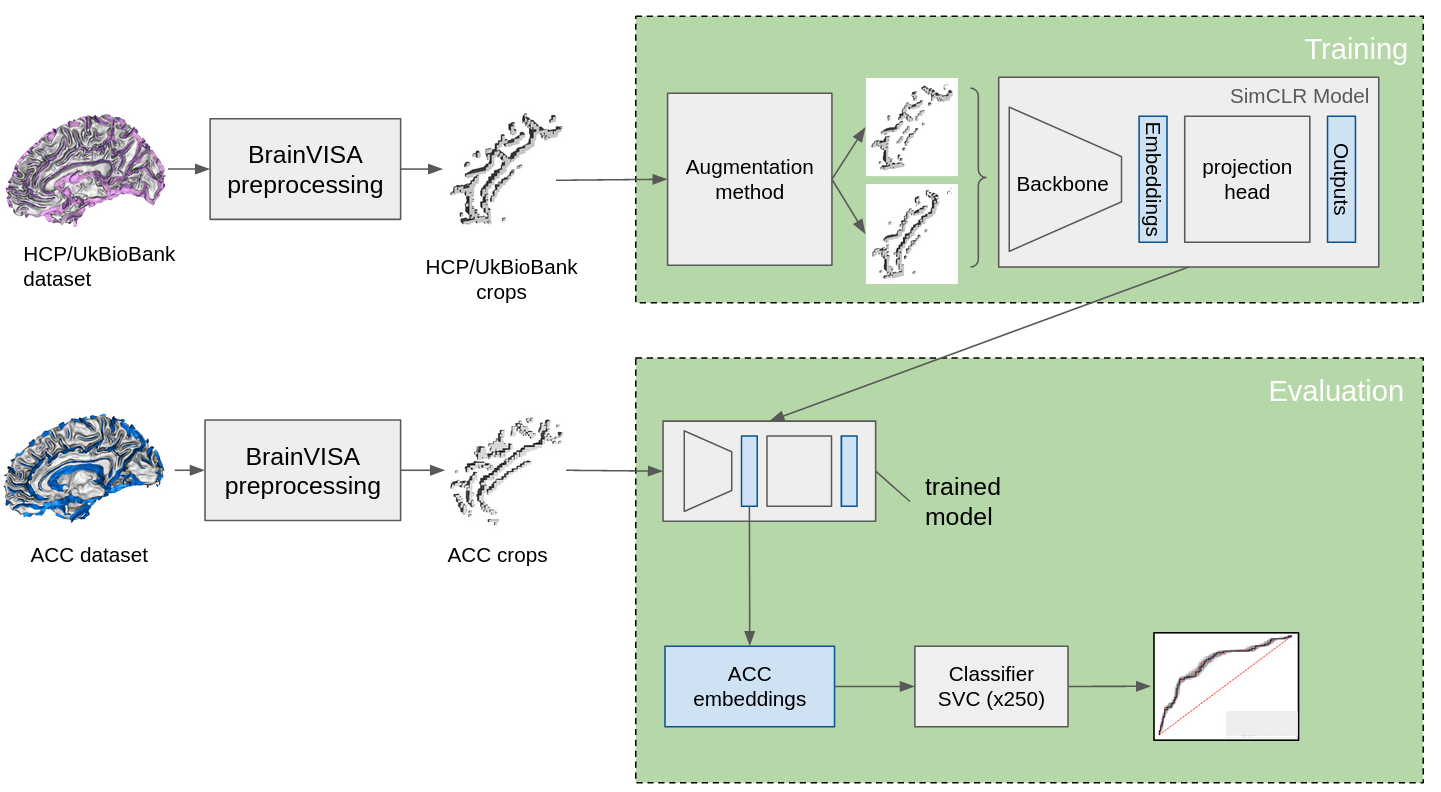}
\caption{\textit{Preprocessing and training pipeline.} We train the SimCLR model either on the HCP or on the UKBioBank dataset (top) and evaluate the model on the ACC dataset, which has been manually labeled for the double-parallel pattern in the cingulate region (bottom).} \label{fig:pipeline}
\end{figure}

\subsection{Deep-learning engineering and parameter optimization}\label{method:optim}

We do initial grid searches using the DenseNet architecture with a 30-dimensional latent space and a non-linear projection head. We are first doing a gridsearch on HCP-1 with evaluation on ACC-1 with a 55\% cutout augmentation~\cite{chavas_unsupervised_2022} over the learning rate on {2, 4}$\times 10^{-4}$, the batch size on {16, 32}. We then set the learning rate to 4$\times 10^{-4}$ and the batch size to 16. We then run a second grid search on {30,45,55}\% cutout size for the cutout augmentation strategy and on {30,40,50}\% of removed pixels for the branch-clipping strategy. We find 55\% for the cutout size and 40\% for the percentage of removed pixels to be the best.

We then test a dropout rate of {0, 0.05, 0.1} for the DenseNet backbone with both a 4-dimension and 30-dimension latent space, with both the cutout and the branch-clipping augmentations. We set the dropout rate to 0.05.

Last, we test the DenseNet backbone when removing or keeping the bottom voxels. For the cutout augmentation, keeping bottom voxels increases the probability of getting stuck in a trivial solution where all outputs lie in the same place, particularly for latent spaces of size 4. For the branch-clipping augmentation, removing bottom voxels improves the AUC, probably by increasing the randomization of the views. Thus, we kept all bottom voxels in the cutout augmentation and removed all bottom voxels in the branch-clipping augmentation.

We then run 5 models for each condition (see~\S\ref{result:optim}). Running a ConvNet model takes 20 minutes on a 16GB GPU (Quadro RT 5000) with 48 CPU cores.

\section{Experiments and results}

\subsection{Optimizing SimCLR model for the detection of the double-parallel pattern}\label{result:optim}

We optimize the SimCLR self-supervised model to detect the double-parallel pattern in the cingulate region. This is done in three main steps: we first perform a preprocessing of the brain MRI data, we then perform parameter optimization on the train/validation dataset, and finally, we test and evaluate the model on the test dataset (Fig.~\ref{fig:pipeline}).

After preprocessing the brain MRI data, we obtain crops of the cingulate region of the right hemisphere containing only information on folds (see~\S\ref{method:preprocessing}). 

\begin{figure}
\centering
\includegraphics[width=0.8\textwidth]{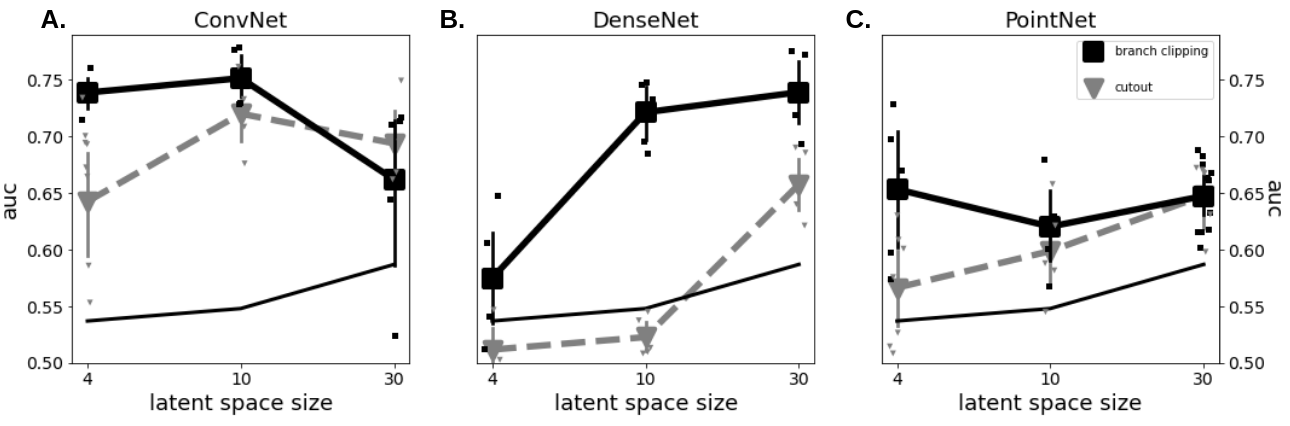}
\caption{\textit{Parameter optimization results.} Training is done on HCP-1 (half of the HCP dataset). Evaluation is done on ACC-1 (half of the ACC dataset). Each small marker represents a different trained model. \textbf{A.} AUC score as a function of the latent space size for the ConvNet backbone. Black lines with squares and dashed lines with down triangles stand respectively for the branch-clipping and the cutout augmentations. The thinner solid black line corresponds to a Principal Component Analysis (PCA) model. \textbf{B.} and \textbf{C.} are similar plots, respectively, for the DenseNet and the PointNet backbones.}
\label{fig:gridsearch}
\end{figure}

We then optimize parameters by training the model on HCP-1 and evaluating it using a linear SVC on ACC-1 (see\S~\ref{method:optim} for details). The evaluation criterion is the AUC for linearly classifying the double-parallel pattern in the cingulate region. After an initial parameter initialization, we test three backbones---a simple six-layer convolutional network (named ConvNet), a DenseNet, and a PointNet---, three latent spaces sizes---4, 10, 30---, two augmentation strategies---the cut-out and the branch-clipping augmentation---, and two projection heads---linear and non-linear---. We find the best model to be the six-layer convolutional network with a linear projection head, a latent space size (that is, the dimension of the latent space) of 10, and using the branch-clipping augmentation strategy (Fig.~\ref{fig:gridsearch}).

\begin{figure}
\centering
\includegraphics[width=0.9\textwidth]{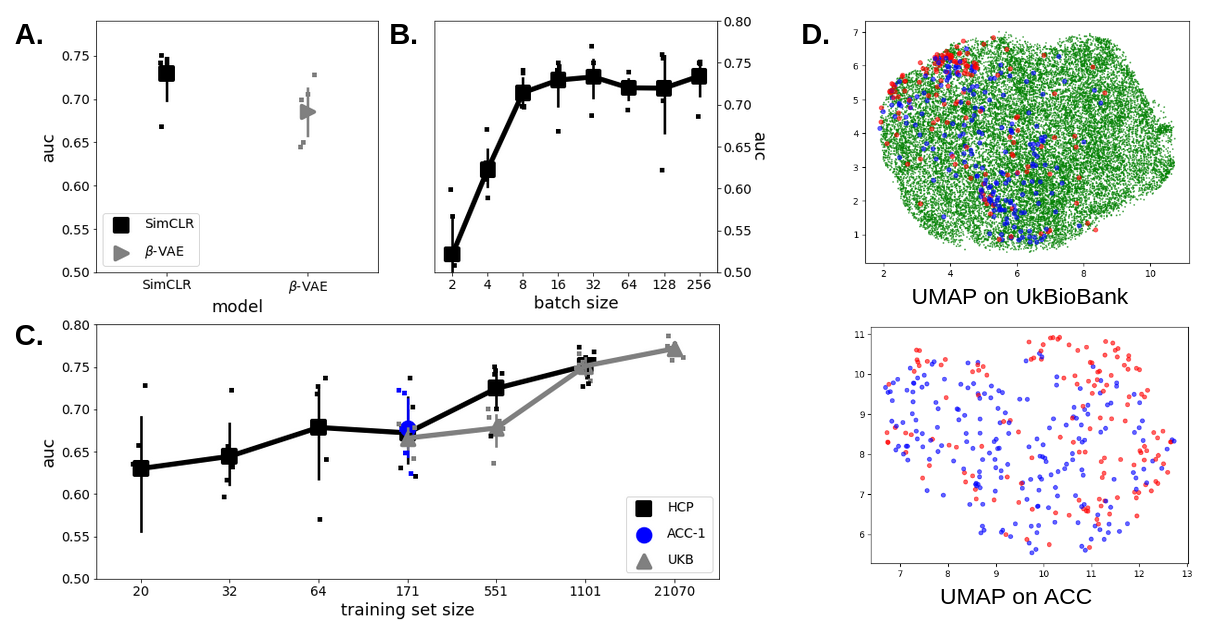}
\caption{\textit{Testing results with the best parameters for SimCLR.} Training is done on HCP-2 and evaluation is done on ACC-2. \textbf{A.} AUC for SimCLR compared with the $\beta$-VAE model. \textbf{B.} AUC as a function of batch size. \textbf{C.} AUC as a function of the training set size. The blue circle and the upper triangle represent models for which the training is done, respectively on ACC-1 and UKBioBank. HCP training is done exclusively on HCP-2 when the training size is smaller than 551. \textbf{D.} UMAP representations of UKBioBank embeddings (top) and of ACC embeddings (bottom). Each point represents a UKBioBank subject (green), an ACC subject with a double-parallel sulcus (red) or without (blue.)} \label{fig:testing}
\end{figure}

Last, for testing, we train models with the chosen hyperparameters on HCP-2 and evaluate them on ACC-2. We find the AUC of SimCLR on the latent space to be 0.73$\pm$0.03 (n=5 models), slightly above the AUC (0.69$\pm$0.04, n=5) of the $\beta$-VAE model optimized in~\cite{chavas_unsupervised_2022} and applied here to the same dataset (Fig.~\ref{fig:testing}A). Even if favorable, we note that the difference doesn't reach statistical significance (p=0.08).

We observe that the effect of the batch size on the AUC reaches a plateau for a small batch size of 8 (Fig.~\ref{fig:testing}B); we use a batch size of 16 in our models after an initial optimization (see~\S\ref{method:optim}). This differs from applying SimCLR on big 2D image datasets, in which bigger batch sizes improve the results~\cite{chen_simple_2020}.
Then, we observe that, by training on the ACC-1 dataset whose size is N=171, the result is similar to the one obtained when training the model on HCP-2 with a similar training size (p=0.84, t-test, comparing blue circle and black square in Fig.~\ref{fig:testing}B). The result is also true for the UKBioBank dataset for enough training data, as the UKBioBank results match the HCP ones for 1101 subjects. As data acquisition sites and age ranges differ in all three datasets, this shows that the method is age and site-resistant. Then, when changing the training set size N, the performances increase. For UKBioBank, the AUC increases to 0.76 when training with 21070 subjects (see Fig.~\ref{fig:testing}C).

Variability (measured as the standard deviation over 5 different SimCLR initializations) decreases from 0.04 for n=551 to 0.01 for n=21070, showing that increasing the number of training subjects significantly stabilizes the representation. Last, the UMAP representation (from a model trained on UKBioBank and fit with UMAP on UKBioBank embeddings) shows how the ACC dataset, as well as the subjects with a double-parallel pattern, is embedded in the UKBioBank space (Fig.~\ref{fig:testing}D).

\section{Discussion and conclusions}

We built the software structure and applied for the first time a self-supervised deep-learning algorithm on cortical skeletons with a database of more than 20,000 subjects. For the first time, we implemented augmentations that use the graph structure of the cortical folds. We optimized the model and quantified its significance using a manually labeled dataset. 

We obtained a latent representation of the cortical folds of the cingulate region and evaluated this representation for detecting a double parallel cingulate pattern known to be related to schizophrenia.

The best SimCLR model found has two characteristics: a very simple backbone---a convolutional neural network---and a topology-specific augmentation---the branch-clipping---that differs from the augmentations used in 2D image learning and takes into account topological information of the inputs. Its simplicity likely means the same model will work reasonably well when applied to other brain regions.

The obtained representation quality is resilient to a change in the training database. This advantage likely comes from using cortical skeletons as they are less sensitive to acquisition sites and the subject's age~\cite{cachia_longitudinal_2016}. However, there is still a dataset effect. A possible solution would be to apply debiasing techniques~\cite{tartaglione_end_2021} or to modify the augmentations further to reduce the remaining site and age effect.

Now that we have a representation that has been optimized for folding pattern detection, the next step will be to bring external information (akin to other modalities) directly into the model to further improve its representation quality~\cite{dufumier_contrastive_2021}. A possible solution may be to perform a late fusion (that is, to concatenate the modalities at the level of the latent space), followed by an autoencoder to leverage possible correlations~\cite{wang_fusing_2022}.

\begin{center}
\textbf{Acknowledgments}
\end{center}
This work was funded by the grants ANR-19-CE45-0022-01IFOPASUBA and ANR-20-CHIA-0027-01FOLDDICO.

\bibliography{main} 
\bibliographystyle{spiebib} 

\end{document}